%% file: main.tex
\documentclass[10pt,twocolumn,letterpaper]{article}

\usepackage[ruled]{algorithm2e} 
\usepackage{booktabs} 
\usepackage{multirow}
\usepackage{booktabs}

\usepackage[pagenumbers]{cvpr} 

\input{preamble}

\definecolor{cvprblue}{rgb}{0.21,0.49,0.74}
\usepackage[pagebackref,breaklinks,colorlinks,citecolor=cvprblue]{hyperref}

\def\paperID{***} 
\def\confName{3DV\xspace}
\def\confYear{2026\xspace}

\title{Gaussian Wardrobe: Compositional 3D Gaussian Avatars\\ for Free-Form Virtual Try-On}
\author{
~~~~~~~~~~~~~~~~~Zhiyi Chen$^{*1}$
\and
Hsuan-I Ho$^{*1}$
\and
Tianjian Jiang$^{1}$~~~~~~~~~~~~~~~~~
\and 
~~~~~~~~~~~~~~~~~~~~~~~~~~~~~Jie Song$^{3,4}$
\and
Manuel Kaufmann$^{1,2}$
\and
Chen Guo$^{\dagger1}$~~~~~~~~~~~~~~~~~~~~~~~~~~~~~
\and
\\
$^{1}$Department of Computer Science, ETH Zürich \\
$^{2}$ETH AI Center, ETH Zürich \\
$^{3}$HKUST(GZ) ~~~~~~~~~~~~~~~~~ $^{4}$HKUST
\and
\url{https://ait.ethz.ch/gaussianwardrobe}
}
\begin{document}
\twocolumn[{%
\renewcommand\twocolumn[1][]{#1}%
\maketitle
\begin{center}
    \centering
    \vspace{-0.6em}
    \captionsetup{type=figure}
    \includegraphics[width=\textwidth]{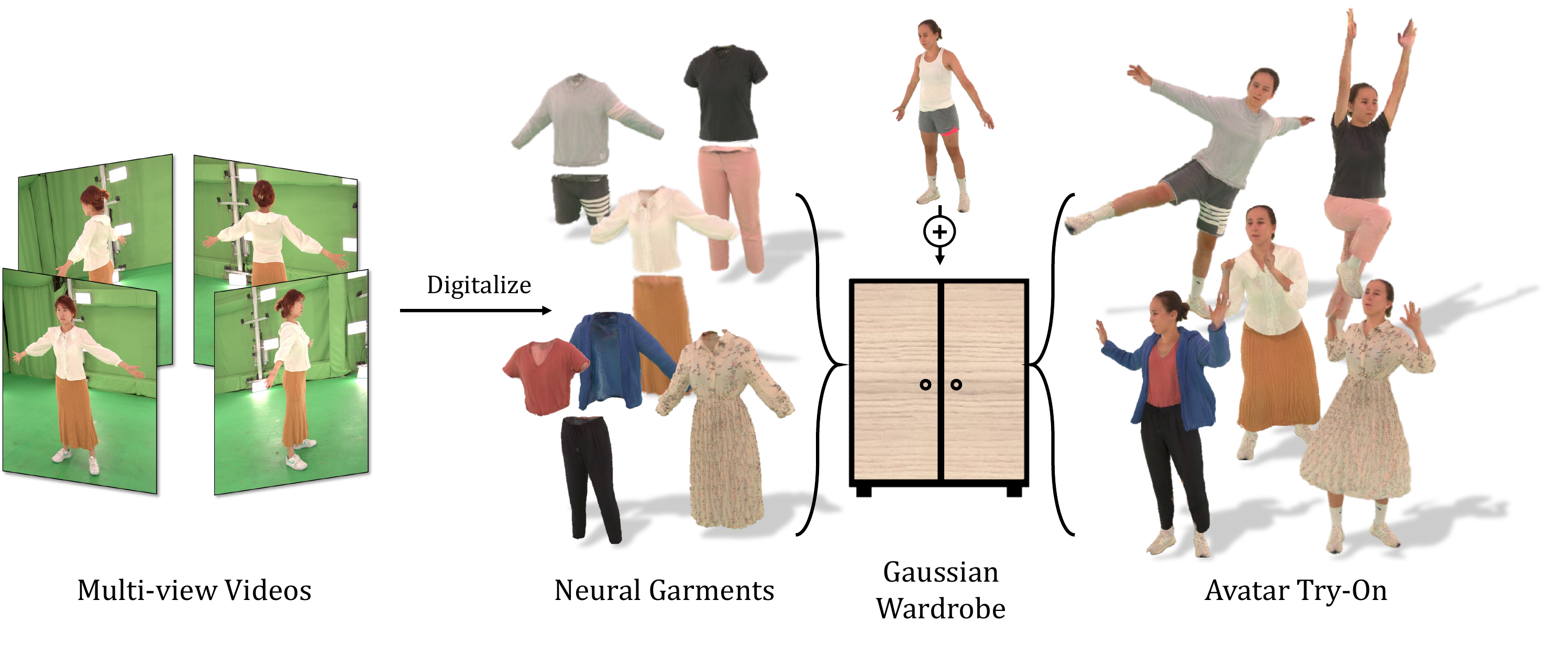}
    \caption{\textbf{Gaussian Wardrobe} is a novel approach to digitalize compositional 3D Gaussian avatars from multi-view videos. The learned neural garments are subject-agnostic. Therefore, they can be stored, reused, and seamlessly recombined with new subjects. Leveraging Gaussian Wardrobe, we realize a practical 3D avatar virtual try-on application. Our method demonstrates the capabilities of modeling the dynamics of challenging free-form clothing such as skirts, dresses, and open jackets. 
    }
    \label{fig:teaser}
\end{center}%
 }]

\def\thefootnote{*}
\footnotetext{Equal contributors\quad\ensuremath{\dagger} Corresponding author}
\def\thefootnote{\arabic{footnote}}

\input{sec/0_abstract}    
\input{sec/1_intro}

\input{sec/2_related_work}
\input{sec/3_method}
\input{sec/4_experiments}
\input{sec/5_conclusion}

{
    \small
    \bibliographystyle{ieeenat_fullname}
    \bibliography{main}
}
\clearpage

\input{supp/6_implement}

\input{supp/7_more_results}

\end{document}

%% file: preamble.tex
\usepackage{makecell}
\usepackage{mathtools}
\usepackage[dvipsnames]{xcolor}

\usepackage{mathtools}

\makeatletter
\renewcommand\paragraph{\@startsection{paragraph}{4}{\z@}%
                {2ex plus 0.2ex minus 0.1ex}%
                {-1.0em}%
                {\reset@font\normalsize\bfseries}}

%% file: sec/0_abstract.tex
\begin{abstract}
We introduce Gaussian Wardrobe, a novel framework to digitalize compositional 3D neural avatars from multi-view videos.
Existing methods for 3D neural avatars typically treat the human body and clothing as an inseparable entity. 
However, this paradigm fails to capture the dynamics of complex free-form garments and limits the reuse of clothing across different individuals.
To overcome these problems, we develop a novel, compositional 3D Gaussian representation to build avatars from multiple layers of free-form garments.
The core of our method is decomposing neural avatars into bodies and layers of shape-agnostic neural garments.
To achieve this, our framework learns to disentangle each garment layer from multi-view videos and canonicalizes it into a shape-independent space.
In experiments, our method models photorealistic avatars with high-fidelity dynamics, achieving new state-of-the-art performance on novel pose synthesis benchmarks.
In addition, we demonstrate that the learned compositional garments contribute to a versatile digital wardrobe, enabling a practical virtual try-on application where clothing can be freely transferred to new subjects.

\end{abstract}

%% file: sec/1_intro.tex
\section{Introduction}
\label{sec:intro}

Recent advancements in neural representations have enabled the creation of photorealistic human avatars with remarkable fidelity~\cite{li2024animatable,guo2025vid2avatarpro}.
These avatars are key across various XR applications, including telepresence, digital fashion, and entertainment. 
However, these methods predominantly train a dedicated neural network to model a clothed human as an indivisible whole.
This one-to-one paradigm limits scalability and prevents avatars for broad user bases.
To overcome this bottleneck, a fundamental property is required for a scalable avatar system: \emph{Compositionality}.

Current neural avatar methods typically represent a clothed human as a single entity~\cite{peng2021animatable, guo2023vid2avatar, jiang2023instantavatar, li2024animatable,moon2024expressive,jiang2025prioravatar}.
While this formulation can efficiently drive the avatars with skeletal deformations from a parametric body model (\eg, SMPL-X~\cite{pavlakos2019expressive}), it strongly assumes that clothing is a simple extension of the body and directly deforms with it.  
This strategy, therefore, fails to capture the dynamics of common garments like open jackets or skirts, which have topologies distinct from the body and do not conform to underlying bone movements.
Some recent attempts~\cite{reloo,scarf,delta,tan2025dressrecon} have begun to incorporate explicit clothing modeling as a distinct entity.
Nevertheless, the underlying deformations of the avatar and its garments remain entangled. This contradicts the real-world nature of clothing, which can be reused, swapped across individuals, and realistically animated on novel human subjects.
The deformation aspect has been largely overlooked and represents the critical factor of \emph{compositionality}.
Ideally, neural avatars should be decomposed into compositional neural garments, defined both in terms of geometry and garment-specific deformation. These garments ought to be subject-agnostic and designed for free recombination across avatars, thereby enabling a versatile digital wardrobe.

In this work, we present \textbf{Gaussian Wardrobe}, a novel framework for digitalizing compositional neural avatars from multi-view videos.
Our method encodes garments of arbitrary topology as animatable 3D Gaussian representations, enabling realistic and dynamic deformations.
The central innovation lies in decomposing a complete human avatar into a collection of subject-agnostic neural garments.
As illustrated in~\cref{fig:teaser}, the garments together form a digital wardrobe that can be recombined and transferred across subjects, thereby facilitating applications like 3D virtual try-on.
The resulting avatars exhibit high-fidelity rendering while faithfully capturing realistic motion dynamics across diverse clothing categories.

At the core of our method is a novel compositional representation designed for garments with arbitrary topologies.
We build upon the recent template-based method of Animatable Gaussians~\cite{li2024animatable}, but introduce three critical modifications to achieve our goals of compositionality.
i) We segment the input template mesh into distinct body and garment components to enable a compositional representation.
This decomposition facilitates more accurate modeling of clothing geometry, appearance, and non-linear motion using 3D Gaussians.
ii) All garments are canonicalized into a zero-shaped space to guarantee cross-subject compositionality.
In this way, the learned appearance and deformation models become inherently shape-agnostic and thus transferable across subjects.
iii) A dedicated learning framework with carefully designed loss functions is developed to disentangle each garment layer from the input videos.
Once trained, these layers are represented as reusable neural assets, collectively forming a digital wardrobe for diverse downstream tasks.

Complementing this reconstruction scheme, we design a practical 3D virtual try-on application with Gaussian Wardrobe. 
Our system leverages the created collection of reusable subject-agnostic neural garments and transfers them to a new user's body shape.
During a try-on session, the target user’s identity is defined by their body Gaussians and shape parameters.
We then combine this identity with the 3D Gaussians of the selected garments and animate the avatar according to a sequence of driving poses.
To mitigate potential surface penetrations under drastically out-of-distribution pose animations, we propose an online penetration detection mechanism integrated into the rendering pipeline.
This strategy enables the detection of interpenetrations between layers and the correction of resulting visual artifacts during rendering.

In our experiments, we demonstrate that the proposed framework creates photorealistic and compositional avatars across diverse garments with complex topologies.
We further establish state-of-the-art performance on the 4D-DRESS~\cite{wang20244ddress} and Actor-HQ~\cite{isik2023humanrf} benchmarks for novel pose synthesis.
In summary, our contributions are threefold:

\begin{compactitem}
    \item A novel compositional 3D Gaussian-based approach capable of modeling 3D human avatars in complex, free-form garments.
    \item A reconstruction scheme that learns to decompose an avatar into distinct neural garments and models their dynamic deformations from multi-view video.
    \item A novel framework to produce subject-agnostic neural garments that enable compositional 3D virtual try-on across subjects with diverse body shapes.
\end{compactitem}

%% file: sec/2_related_work.tex
\section{Related Work}
\paragraph{Animatable 3D human avatar.}

Traditionally, 3D human avatars were created using pre-scanned human meshes~\cite{monoperfcap,livecap,deepcap,deepdychar} or extensions of parametric body models~\cite{Zhang_2017_CVPR,Alldieck_2018_CVPR,joo2018total,tiwari20sizer,alldieck2021imghum}.
However, these approaches often suffer from limited representational capability.
To address this issue, the research community introduced neural representations, such as implicit neural fields~\cite{Mescheder_2019_CVPR,Park_2019_CVPR}, Neural Radiance Fields (NeRF)~\cite{mildenhall2020nerf}, and 3D Gaussians~\cite{3dgs}, to model clothed humans as neural avatars.
Numerous works have since been proposed to digitize these avatars from 3D scans~\cite{Chen2023fastsnarf, shen2023xavatar, li2024animatable}, monocular videos~\cite{peng2021animatable,guo2023vid2avatar,jiang2023instantavatar,neuman,hu2024gaussianavatar,hu2024gauhuman,moon2024expressive,jiang2025prioravatar}, and even single images~\cite{saito2019pifu,xiu2022icon,xiu2023econ,zhang2024sifu,ho2024sith}.
While these methods produce high-quality appearances, they represent the body and clothing as a single entity, which limits their expressiveness for modeling free-flowing garments.

Recently, another stream of work has emerged that explicitly models garments as a separate entity. 
These approaches extend the underlying parametric body models with additional implicit~\cite{jiang2020bcnet, corona2021smplicit, moon20223d, scarf, delta, reloo,vuran2025remu} or explicit representations~\cite{ma2021scale, ma2021power, ma2022skirt,lin2022learning,zhang2023closet,prokudin2023dynamic,tan2025dressrecon, zielonka25dega} to capture the complex surface deformations of clothing.
However, a fundamental limitation is that the deformations of the avatar and its garments remain entangled, thereby restricting their applicability in realistic virtual try-on settings.
While some attempts incorporate physically-simulated training data or physics-informed objectives~\cite{zheng2024physavatar, su2023caphy, li2024idffavatar, rong2024gaussiangarments, santesteban2022snug, grigorev2023hood}, they typically focus on modeling the geometric deformation of a single garment and do not support holistic animation of the human body together with garments comprising multiple layers.
The work most closely related to ours is LayGA~\cite{layga}, which models multi-layer garments using 3D Gaussians derived from an underlying SMPL-X body mesh.
Although this approach enables compositional virtual try-on, its reliance on parametric body meshes restricts its ability to represent free-form loose clothing, such as open jackets and skirts, whose dynamics are only weakly correlated with skeletal deformations.
In this work, we address this limitation by integrating mesh-based garment modeling with state-of-the-art animatable 3D Gaussian representations.
This integration enables our method not only to support novel clothing transfer for tight-fitting garments but also to achieve compelling results when dressing subjects in loose outfits.

\paragraph{Avatar virtual try-on and editing.}
The literature on virtual try-on has mostly focused on composing outfits within 2D images~\cite{han2018viton,sun2024outfitanyone, song2025image}.
With recent developments in XR technology, 3D virtual try-on has emerged as a significant research area. 
Conventionally, 3D avatar customization was achieved by combining extensive collections of human scans with artist-designed 3D assets~\cite{Black_CVPR_2023, zou2023cloth4d}.
To avoid this time-consuming process, recent approaches~\cite{instructnerf2023,zhu2024InstructHumans,cao2024gsvton,pang2025disco} have leveraged Score Distillation Sampling (SDS)~\cite{poole2023dreamfusion} to edit avatar appearance and clothing using text prompts.
While these techniques offer flexibility, their lengthy optimization times limit their practicality.

More closely related to our work are methods that reconstruct interchangeable 3D avatars for virtual try-on and editing~\cite{ho2023custom,layga,zhang2025layeravatar}.
For instance, CustomHumans~\cite{ho2023custom} learns a generative model over SMPL-X body meshes, enabling the swapping of local appearances at each mesh face.
Similarly, LayGA~\cite{layga} uses SMPL-X meshes to predict 3D Gaussians for garments, while LayerAvatar~\cite{zhang2025layeravatar} employs diffusion models to generate clothing textures and displacements from the UV maps of parametric bodies.
A significant limitation of these methods is that their compositional ability stems from the shared topology of the underlying parametric mesh.
This dependence restricts them to modeling only a few garment layers and fails to capture the free-form dynamics of real-world clothing. 
In contrast, Gaussian Wardrobe is inspired by mesh-based neural avatars~\cite{li2024animatable,rong2024gaussiangarments}, which offer greater capacity to model diverse garments and achieve a more realistic virtual try-on experience.

%% file: sec/3_method.tex
\input{sec/FIG_method}

\section{Gaussian Wardrobe}

\paragraph{Preliminary.} Our compositional Gaussian Avatar is built upon the framework of Animatable Gaussians~\cite{li2024animatable}.
Given a multi-view video $\{I_{m,n}\}$ of a human subject, where $m$ is the frame index and $n$ is the camera index, we register the subject's shape and per-frame body pose parameters.
We use the SMPL-X~\cite{pavlakos2019expressive} parametric body model, which defines shape with $\beta \in \mathbb{R}^{10}$ PCA coefficients and pose with parameters $\theta \in \mathbb{R}^{21+30+3+1}$
(comprising 21 body joint rotations, 30 hand parameters, 3 facial joint rotations, and a global orientation).

The Animatable Gaussians method utilizes a mesh template and a pose-conditioned U-Net to represent the avatar.
To create the template, we first select the first video frame $\{I_{1,n}\}$, which should feature a pose close to the standard A-pose.
We then reconstruct a mesh from this frame using the multi-view data and repose it into a pre-defined canonical pose.
This template, denoted as $\mathcal{M}$ in~\cref{fig:arch} (left), is used to render pose-dependent positional maps for any given body pose $\theta_m$.
The U-Net is trained to predict the parameters of 3D Gaussian primitives for various input poses, which enables the avatar to be driven by novel movements.

\paragraph{Method overview.} 
The overall framework of our method, Gaussian Wardrobe, is summarized in \cref{fig:arch}.
In \cref{sec:rep}, we present our approach for augmenting the Animatable Gaussians representation to support compositionality and model free-form neural garments. 
Next, in \cref{sec:train}, we introduce the learning framework used to decompose individual garment layers from multi-view videos.
Finally, in \cref{sec:tryon}, we demonstrate a practical application of 3D virtual try-on using a collection of trained neural garments that form a digital wardrobe.

\subsection{Compositional Gaussian Representation}
\label{sec:rep}

A single template mesh is insufficient for modeling the complex dynamics of free-form clothing. 
Moreover, because the template encodes subject-specific shape information ($\beta$), it is not suitable for generalization across different subjects. 
To address these limitations, we design a new compositional Gaussian representation.

First, to make the template shape-agnostic, we deform it into a canonical, zero-shape space by removing the subject-specific shape blendshapes $\beta$.  Specifically, each vertex $\mathbf{v}^{t}_i$ on the original template is displaced to a new position $\mathbf{v}^{c}_i$ following: 
$\mathbf{v}^c_i = \mathbf{v}^{t}_i - \sum_{b=1}^{10} \beta_b \mathbf{o}_{bi}.$
Here, $\beta_b$ is the $b$-th PCA shape coefficient, and $\mathbf{o}_{bi}$ is the corresponding blendshape offset for the $i$-th vertex.
To obtain these offsets for our custom template $\mathcal{M}$, we adopt the strategy from Fast-SNARF~\cite{Chen2023fastsnarf}, which involves voxelizing and diffusing the SMPL-X blendshape offsets into a $64^3$ voxel grid.
The value of $\mathbf{o}_{bi}$ for each vertex is then queried by trilinearly interpolating the values in this grid at the position of $\mathbf{v}^{t}_i$.

After obtaining the shape-agnostic template, we perform 3D segmentation~\cite{wang20244ddress} to separate the body from the different garment layers. 
Specifically, we segment the template into upper ($\mathcal{M}_u$), lower  ($\mathcal{M}_\ell$), and optional outer $(\mathcal{M}_o$) garment layers to model their dynamic deformations.
To capture the underlying skeletal motion, we use a separate SMPL-X body mesh ($\mathcal{M}_b$).
The meshes for regions like hair and shoes are merged with this body mesh to better preserve their details.
We visualize these multi-layer, shape-agnostic templates in~\cref{fig:arch} (left).

For each layer $L \in {\{b, u, \ell\}}$, we follow the approach of Animatable Gaussians~\cite{li2024animatable} to generate two feature maps from its corresponding template mesh $\mathcal{M}_L$.
First, we create front-and-back coordinate maps $\mathbf{C}_L \in \mathbb{R}^{2\times H\times W\times3}$ by rasterizing the template from orthographic front and back views.
Each pixel in these maps stores the 3D position of a point on the template, defining the initial state of the 3D Gaussians.
Second, we deform the template mesh using Linear Blend Skinning (LBS) with the registered pose parameters $\theta_m$ to generate posed positional maps $\mathbf{P}_L(\theta_m) \in \mathbb{R}^{2\times H\times W\times3}$.
These maps have the same dimensions as the coordinate maps, but each pixel stores the new 3D position of the corresponding point after deformation.

These positional maps serve as input to a set of layer-specific neural networks, $\mathcal{F}_L$.
Each network is trained to predict a corresponding front-and-back Gaussian map $\mathbf{M}_L \in \mathbb{R}^{2\times 2H\times 2H\times (14)}$ which represents the clothing or body part in the target pose $\theta_m$: 
\begin{equation}
\mathcal{M}_{L}
\xmapsto{}
\big(\mathbf{C}_L,\mathbf{P}_L(\theta_m) \big)
\xmapsto{\mathcal{F}_L}
\mathbf{M}_{L}, ~~~~L \in {\{b, u, \ell\}}.
\label{eq:unet}
\end{equation}
Each pixel in the output Gaussian map $\mathbf{M}_L$ defines a 3D Gaussian primitive with parameters: positional offset, rotation, opacity, scale, and color $[\Delta \mathbf{p}^c_i, \mathbf{q}_i, \alpha_i, \mathbf{s}_i, \mathbf{c}_i] \in \mathbb{R}^{14}$.
The associated covariance matrix is calculated as $\mathbf{\Sigma}_i=\mathbf{RSS^{\intercal}R^{\intercal}}$, where $\mathbf{R}$ is the rotation matrix derived from $\mathbf{q}_i$, and $\mathbf{S}$ is the diagonal scaling matrix derived from $\mathbf{s}_i$.

\subsection{Learning Compositional Neural Garments}
\label{sec:train}
Given multi-view video frames $\{I_{m,n}\}$ and their corresponding body poses $\theta_m$, our goal is to train a separate neural network for each garment layer to model its dynamics.
To do so, the predicted shape-agnostic Gaussian primitives must be deformed back into the target posed space.
This allows them to be rendered and compared against the ground-truth 2D images for computing training losses.

For a Gaussian primitive from any layer $L \in {\{b, u, \ell\}}$, we start with its canonical position $\mathbf{p}^c_i$ from the coordinate map $\mathbf{C}_L$.
We first apply the network's predicted positional offset $\Delta\mathbf{p}^c_i$.
Next, we transform this updated point from the canonical space into the final posed space by reapplying the subject-specific shape blendshapes $\beta$ and performing LBS with respect to the pose $\theta_m$.
This process updates the position and rotation of each Gaussian primitive as follows:

\begin{equation}
\begin{aligned}
    \hat{\mathbf{p}}_i &= \sum_{j=1}^{55}w_{ij}\big(\mathbf{R}_{j} (\mathbf{p}_i^c + \Delta\mathbf{p}^c_i + \sum_{b=1}^{10} \beta_b \mathbf{o}_{bi} )  + \mathbf{t}_j \big), \\
    \hat{\mathbf{\Sigma}}_i &= \sum_{j=1}^{55} w_{ij} (\mathbf{R}_j \mathbf{\Sigma}_i \mathbf{R}^{\intercal}_j). 
\label{eq:deform}
\end{aligned}
\end{equation}
Here, ($\mathbf{R}_j,\mathbf{t}_j$) is the $j$-th bone transformation mapping from the canonical pose to the target pose $\theta_m$, and $w_{ij}$ are the skinning weights obtained via the voxelized field strategy discussed in \cref{sec:rep}.

After this deformation, we obtain a set of renderable 3D Gaussians for each layer $\mathcal{G}_L$ correctly positioned in the posed space.
As shown in \cref{fig:arch} (right), we composite the Gaussians from all layers and use a splatting-based rasterizer $\mathcal{R}_{splat}(\cdot)$ to render RGB images ($\hat{I}_{m,n}$) and segmentation masks $\hat{S}_{m,n}$ from the $n$-th camera viewpoint $\mathbf{\Pi}_{n}$:

\begin{equation}
\big(\mathcal{G}_{b},\, \mathcal{G}_{u},\, \mathcal{G}_{\ell}\big)
\xmapsto{\mathcal{R}_{\mathrm{splat}}(\mathbf{\Pi}_{n})}
\big(\hat{I}_{m,n},\, \hat{S}_{m,n} \big).
\label{eq:render}
\end{equation}
Finally, we jointly optimize all layer-specific U-Nets using the objective functions described next.

\paragraph{Photometric losses.}
Similar to standard 3D Gaussian Splatting, we enforce photometric loss between the rendered images ($\hat{I}_{m,n}$) and the ground-truth video frames  ($I_{m,n}$). The loss is a combination of $L_1$ loss, SSIM~\cite{wang2003multiscale} loss, and LPIPS~\cite{zhang2018perceptual} loss terms:

\begin{equation}
\begin{aligned}
\mathcal{L}_{\text{im}} = & \sum_{m,n} \big( \|\hat{I}_{m,n} - I_{m,n}\|_1 + \\
& \lambda_{1}\mathcal{L}_{\text{SSIM}} (\hat{I}_{m,n}, I_{m,n}) + \lambda_{2}\mathcal{L}_{\text{Lpips}}(\hat{I}_{m,n}, I_{m,n}) \big ).
\end{aligned}
\end{equation}
To improve facial detail, we extract the face regions ($F$) from both the rendered and ground-truth images and compute an additional face-specific LPIPS loss:
\begin{equation}
\begin{aligned}
\mathcal{L}_{\text{f}} = & \sum_{m,n} \lambda_{f}\mathcal{L}_{\text{Lpips}}(\hat{F}_{m,n}, F_{m,n}).
\end{aligned}
\end{equation}

\paragraph{Segmentation losses.}
A primary objective of our method is to ensure that each garment layer is correctly decomposed and separated.
To enforce this, we employ a segmentation loss inspired by D3GA~\cite{zielonka25dega}.
We first assign a distinct color to the Gaussians from each layer and render multi-class segmentation masks $\hat{S}_{m,n}$ (visualized in \cref{fig:arch} (right)).
These are compared against ground-truth masks $S_{m,n}$ obtained from an off-the-shelf segmentation model~\cite{wang20244ddress}.

Furthermore, since the underlying body has minimal pose-dependent dynamics, its shape should remain close to the registered template.
We enforce this by rendering body-only masks, $\hat{S}^b_{m,n} = \mathcal{R}_{\mathrm{splat}}(\mathcal{G}_{b};\mathbf{\Pi}_{n})$, and comparing them to masks rendered from the input body template $S^b_{m,n}$.
The total segmentation loss is:

\begin{equation}
    \mathcal{L}_{\text{sg}} = \sum_{m,n} \big( \lambda_{\text{sg}}\|\hat{S}_{m,n} - S_{m,n}\|_2 + \lambda_{\text{bs}}\mathcal{L} \|\hat{S}^b_{m,n}- S^b_{m,n}\|_1 \big)
\end{equation}

\input{sec/FIG_pene}
\paragraph{Regularization.}
We include two types of regularization to ensure the reconstructed neural body and garments are physically plausible.

First, to prevent inner layers from penetrating outer layers, we introduce a penetration loss.
For a Gaussian at position $\mathbf{p}_u$ on an outer layer (e.g., a shirt), we find its nearest neighbor at position $\mathbf{p}_b$ on the adjacent inner layer (e.g., the body).
We then enforce that the signed distance between them along the inner layer's normal vector $\mathbf{n}_b$ should be larger than a minimum threshold $\epsilon$.
This is formulated as a squared hinge loss:

\begin{equation}
    \mathcal{L}_{\text{pe}} = \max(\epsilon - (\mathbf{p}_u -\mathbf{p}_b)\cdot \mathbf{n}_b, 0)^2.
    \label{eq:penetration}
\end{equation}
Note that this regularization is applied between all adjacent layers. Please see the supplementary material for details on estimating the normal vectors $\mathbf{n}_b$.

Second, we apply several geometric regularizations for stable convergence.
These include an offset term $\mathcal{L}_{\text{o}} = ||\Delta\mathbf{p}||^2_2$ to penalize large Gaussian displacements;
a smoothing term $\mathcal{L}_{\text{sm}} = ||\Delta\mathbf{p}-{\Delta\mathbf{p}_\mathcal{N}} ||^2_2 $ to ensure coherent offsets among neighboring Gaussians;
and a body opacity term $\mathcal{L}_{\text{bo}} = - \log(\alpha_b)$ to encourage the body layer to be fully opaque. The regularization terms are combined into a single loss:
\begin{equation}
\mathcal{L}_{\text{reg}} = \lambda_{\text{pe}} \mathcal{L}_{\text{pe}} + \lambda_{\text{o}} \mathcal{L}_{\text{o}} + \lambda_{\text{sm}} \mathcal{L}_{\text{sm}} + \lambda_{\text{bo}} \mathcal{L}_{\text{bo}}.
\end{equation}

\paragraph{Full Objective.}
Finally, we jointly optimize all U-Nets by minimizing the full loss function, which is a weighted sum of all components:

\begin{equation}
\mathcal{L}_{\text{full}} = \mathcal{L}_{\text{im}} + \mathcal{L}_{\text{f}} + \mathcal{L}_{\text{sg}} + \mathcal{L}_{\text{reg}}.
\end{equation}
The $\lambda_{(\cdot)}$ terms are hyperparameters used to balance each loss component. Their specific values are listed in the supplementary material.

\subsection{Avatar Virtual Try-on}
\label{sec:tryon}

Once trained, each shape-agnostic garment layer can be stored, reused, and seamlessly swapped across multiple human subjects.
We exploit this compositional nature to develop a practical virtual try-on application.

In this application, the user's identity is defined by their body shape parameters  $\beta^*$ and the corresponding body template and U-Net $(\mathcal{M}_b^*,\mathcal{F}_b^*)$.
This ensures that their unique skin tone, hairstyle, and face remain consistent during a try-on session. 
With the body fixed, we can replace any garment layer's template and network, such as $(\mathcal{M}_{\{u,\ell,o\}}^*, \mathcal{F}_{\{u,\ell,o\}}^*)$ with other clothing items from our digital wardrobe.
For example, a user can swap shorts for a skirt by replacing $(\mathcal{M}_\ell^*,\mathcal{F}_\ell^*)$ with a new pair $(\mathcal{M}_\ell',\mathcal{F}_\ell')$ as shown in~\cref{fig:penetration} (left).
Finally, we composite a new set of 3D Gaussians for the complete outfit and render a final image $\hat{I}'$.
This step follows the deformation process in~\cref{eq:deform} but uses the user's shape $\beta^*$, a novel driving pose$\theta'$, and a target camera matrix $\mathbf{\Pi}'$:

\begin{equation}
\big(\mathcal{M}_{b}^*,\, \mathcal{M}_{u}^*,\, \mathcal{M}_{\ell}'\big)
\xmapsto{(\theta', \beta^*)}
\big(\mathcal{G}_{b}^*,\, \mathcal{G}_{u}^*,\, \mathcal{G}_{\ell}'\big)
\xmapsto{\mathcal{R}_{\mathrm{splat}}(\mathbf{\Pi}')}
\hat{I}'.
\end{equation}

\paragraph{Penetration-aware Rendering.}
\label{sec:penetration-aware-rendering}
While our penetration regularizer effectively suppresses penetration during training, minor visual artifacts may still persist during animation, particularly with out-of-distribution poses (see \cref{fig:penetration} (middle)).
To resolve this, we developed a simple yet efficient online correction algorithm that integrates directly into the rendering pipeline.
During inference, alongside the RGB image, we render multi-class segmentation masks $\hat{S}'$ for the composed outfit.
We then apply a contour-finding algorithm \cite{SUZUKI198532} to these masks to detect discontinuous regions, which indicate potential inter-garment penetration.
By analyzing the depth values of the Gaussians within these detected regions, we can confirm if a pixel from an inner layer was rendered on top of an outer layer.
If penetration is confirmed, we correct the erroneously rendered pixel by replacing its color with that of the correct outermost garment.
As shown in~\cref{fig:penetration} (right), this post-processing step effectively removes visual artifacts.
This algorithm is efficient for correcting penetrations on the fly.

It is worth noting that we do not explicitly handle inter-layer penetration like previous methods~\cite{layga,vuran2025remu}, since our goal is to produce realistic 2D rendering rather than to solve the correct 3D geometry. With our online correction strategy, we can mitigate the expensive geometry optimization process and simplify the problem to selecting among the rendering layers more efficiently.

%% file: sec/FIG_method.tex
\begin{figure*}[!ht]
    \centering
    \captionsetup{type=figure}%
    \includegraphics[width=\linewidth]{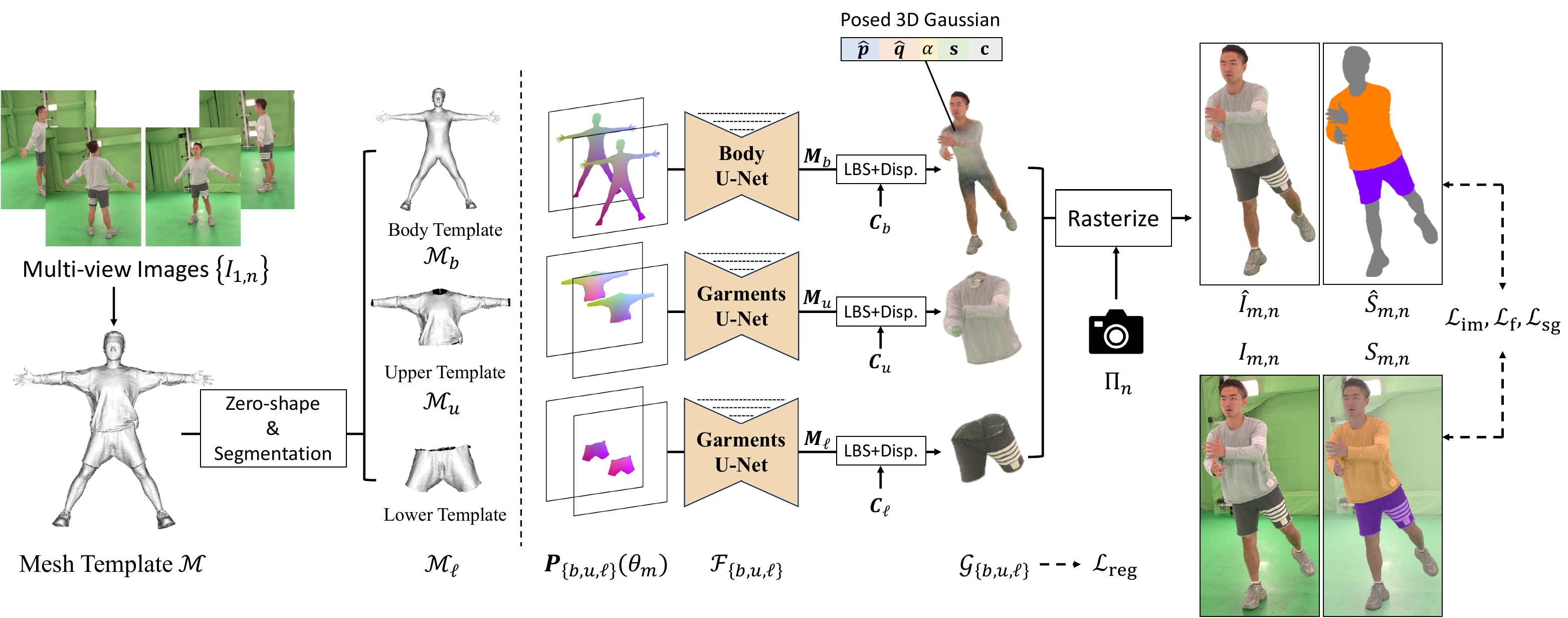}
    \caption{\textbf{Gaussian Wardrobe} digitalizes compositional neural avatars from multi-view videos.
    Our pipeline consists of two major components: (left) a compositional Gaussian representation and (right) a framework for learning neural garments. We first reconstruct a mesh template $\mathcal{M}$ from the first video frame and segment it into body $\mathcal{M}_b$, garment templates $\mathcal{M}_{\{u,\ell\}}$ in the zero-shaped canonical space. During training, each layer learns a separate U-Net $\mathcal{F}$ to predict the parameters of 3D Gaussian primitives $\mathbf{M}$ from pose-conditioned positional maps $\mathbf{P}$. We composite the 3D Gaussians $\mathcal{G}$ from all layers and render RGB images $\hat{I}$ and segmentation masks $\hat{S}$ to compute the training loss $\mathcal{L}$. The learned neural garments are shape-agnostic and can seamlessly transfer to other subjects for avatar virtual try-on.
    }
    \label{fig:arch}
\end{figure*}

%% file: sec/FIG_pene.tex
\begin{figure}[t]
    \centering
    \includegraphics[width=\linewidth]{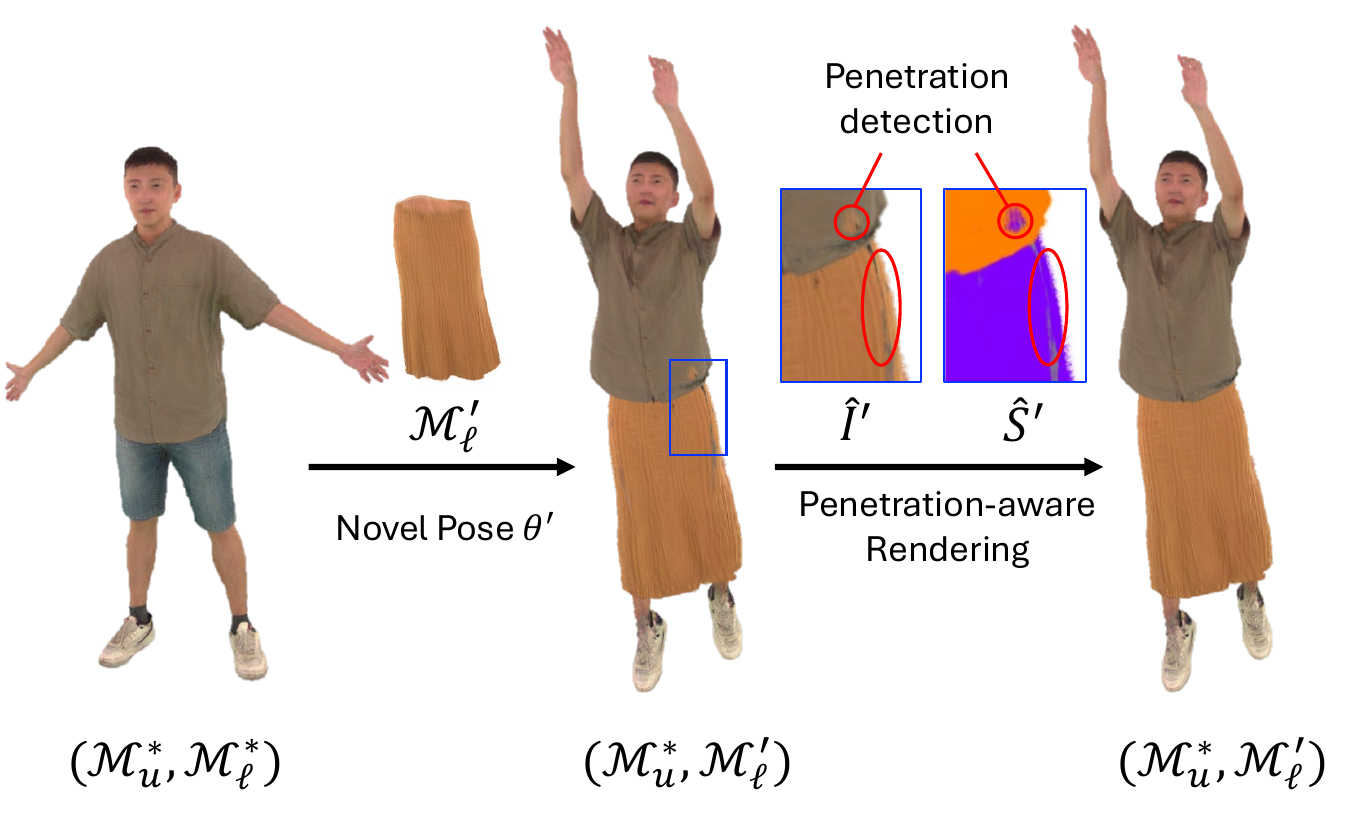}
    \caption{ \textbf{Exemplar of avatar virtual try-on.} Given a reconstructed avatar (left), we replace its lower garment with a new skirt, $\mathcal{M}_\ell'$. The combined avatar can be animated to a novel pose $\theta'$, which may introduce minor penetration artifacts (middle). We resolve these artifacts on-the-fly during rendering with our online penetration detection algorithm (right).
    }
    \label{fig:penetration}
\end{figure}

%% file: sec/4_experiments.tex
\input{tab/TAB_recon_full_vertical}

\input{sec/FIG_res_deformation}

\section{Experiments}
\subsection{Experiment Setup}
\paragraph{Datasets.} 
We evaluate our method on two public datasets: 4D-DRESS~\cite{wang20244ddress} and ActorsHQ~\cite{isik2023humanrf}.
\begin{itemize}
    \item 4D-DRESS contains multi-view videos of 32 subjects in diverse outfits and challenging poses.
    For our method, we select 7 subjects (16 distinct garments in total) and use 700-1100 frames from 12 camera views for training, holding out 300 frames per subject for evaluation.
    Due to the significant training time of baseline methods, we retrain them on a representative subset of 3 of these subjects for comparison.
    \item ActorsHQ features multi-view videos of 8 subjects performing simpler body poses.
    From this dataset, we utilize 3 subjects, with approximately 1000 training frames and 350 test frames captured from 14 views. All methods were trained on the same data for this dataset.
\end{itemize}

\paragraph{Evaluation Protocol.}
For the quantitative evaluation of novel pose synthesis, we report three standard metrics: PSNR, SSIM~\cite{wang2003multiscale}, and LPIPS~\cite{zhang2018perceptual}.
Additionally, to specifically assess the quality of our garment decomposition in the ablation studies, we employ segmentation metrics including mean mIoU, Recall, and F1-Score.

\subsection{Novel Pose Synthesis Comparisons}
\label{sec:exp_pose}
We compare our method against two state-of-the-art baselines: Animatable Gaussians~\cite{li2024animatable} and LayGA~\cite{layga}.
For Animatable Gaussians, we use the official public implementation and adhere to the default training configurations.
As the source code for LayGA is not public, we re-implemented the method based on direct guidance from the original authors to ensure fidelity.
To ensure a fair comparison, all methods are evaluated on the identical, held-out test sets corresponding to the subjects on which they were trained.

The quantitative results in~\cref{tab:novel-pose-all}, demonstrate that our method outperforms both baselines across all metrics on both datasets. 
Qualitative comparisons, shown in~\cref{fig:deform_compare}, further highlight the strengths of our approach.
Our method better models the complex dynamics of free-form garments, such as the flowing motion of a skirt or the flaps of a vest, where SMPL-X mesh-based methods (i.e., LayGA) often fail.
Furthermore, our approach captures finer details, particularly in facial regions and along clothing boundaries.
We attribute this improvement to our explicit decomposition of the avatar, which allows the model to learn dedicated representations for the body and each garment layer.
This addresses the blurred or entangled artifacts common in unified models like Animatable Gaussians. For more qualitative comparisons on free-form garments, please refer to the supplementary video.

\subsection{Ablation Studies}
We conduct a series of ablation studies to validate the loss functions designed for our learning framework.
In each experiment, we remove a specific loss term during training and evaluate its impact on garment decomposition using segmentation metrics.
The quantitative and qualitative results are summarized in~\cref{tab:ablation-regularizers-S} and~\cref{fig:ablation-penetration}, respectively.

\input{tab/ablation}
\input{sec/FIG_ablation}

First, we examine the effect of the segmentation loss $\mathcal{L}_{sg}$. Without this term, the body and clothing representations become significantly entangled.
As visualized during virtual try-on in~\cref{fig:seg}, this entanglement prevents successful garment transfer, making the garment's appearance corrupted when applied to a new subject.
Furthermore, the segmentation loss alone is insufficient to prevent physical artifacts like inter-layer penetration between the body and clothing.
This is evident in the variants that lack our regularization $\mathcal{L}_{reg}$ and penetration $\mathcal{L}_{pe}$ losses, where visible penetration artifacts occur.
By combining all proposed loss terms, our full model achieves a clean, penetration-free decomposition.
These results validated the necessity of each component.
Please refer to the supplementary material for more results.

\subsection{Virtual Try-On Applications}
A key application of our framework is a flexible 3D virtual try-on system.
Because our learned garments are subject-agnostic, they can be reused and transferred across different subjects. 
As demonstrated in~\cref{fig:clothing_transfer}, these garments seamlessly adapt to new subjects with varying body shapes and identities.
It is worth noting that the resulting avatars are not static; they can be animated by novel pose sequences and exhibit realistic, free-flowing dynamics in unseen poses.
Please refer to our supplementary video for more demonstrations of these dynamic try-on results.

\input{sec/FIG_segmentation}
\input{sec/FIG_virtual_tryon}

%% file: tab/TAB_recon_full_vertical.tex
\begin{table}[t]
    \centering
    \small
\begin{tabular}{l|ccc}
\toprule
Method               & PSNR $\uparrow$ & SSIM $\uparrow$ & LPIPS $\downarrow$  \\
\midrule
                     \multicolumn{4}{c}{4D-DRESS~\cite{wang20244ddress}}  \\
\midrule

Animatable Gaussians~\cite{li2024animatable} & 27.75 & 0.9577 & 0.0531  \\
LayGA~\cite{layga} & 27.58 & 0.9574 & 0.0543  \\
Gaussian Wardrobe (Ours) & \textbf{28.06} & \textbf{0.9579} & \textbf{0.0527} \\
\midrule
                      \multicolumn{4}{c}{ActorsHQ~\cite{isik2023humanrf}}  \\
\midrule
Animatable Gaussians~\cite{li2024animatable} & 27.92 & 0.9411 & 0.0418  \\
LayGA~\cite{layga} & 27.80 & 0.9413 & 0.0421  \\
Gaussian Wardrobe (Ours) & \textbf{28.38} & \textbf{0.9436} & \textbf{0.0397} \\
 
\bottomrule
\end{tabular}

\caption{\textbf{Evaluation of novel pose synthesis on 4D-DRESS and ActorsHQ datasets.} Best results are highlighted in \textbf{bold}. Our method consistently outperforms all baselines on all evaluation metrics (\cf~\cref{fig:deform_compare}).}

\label{tab:novel-pose-all}
\end{table}

%% file: sec/FIG_res_deformation.tex
\begin{figure*}[t]
    \centering
    \includegraphics[width=\linewidth]{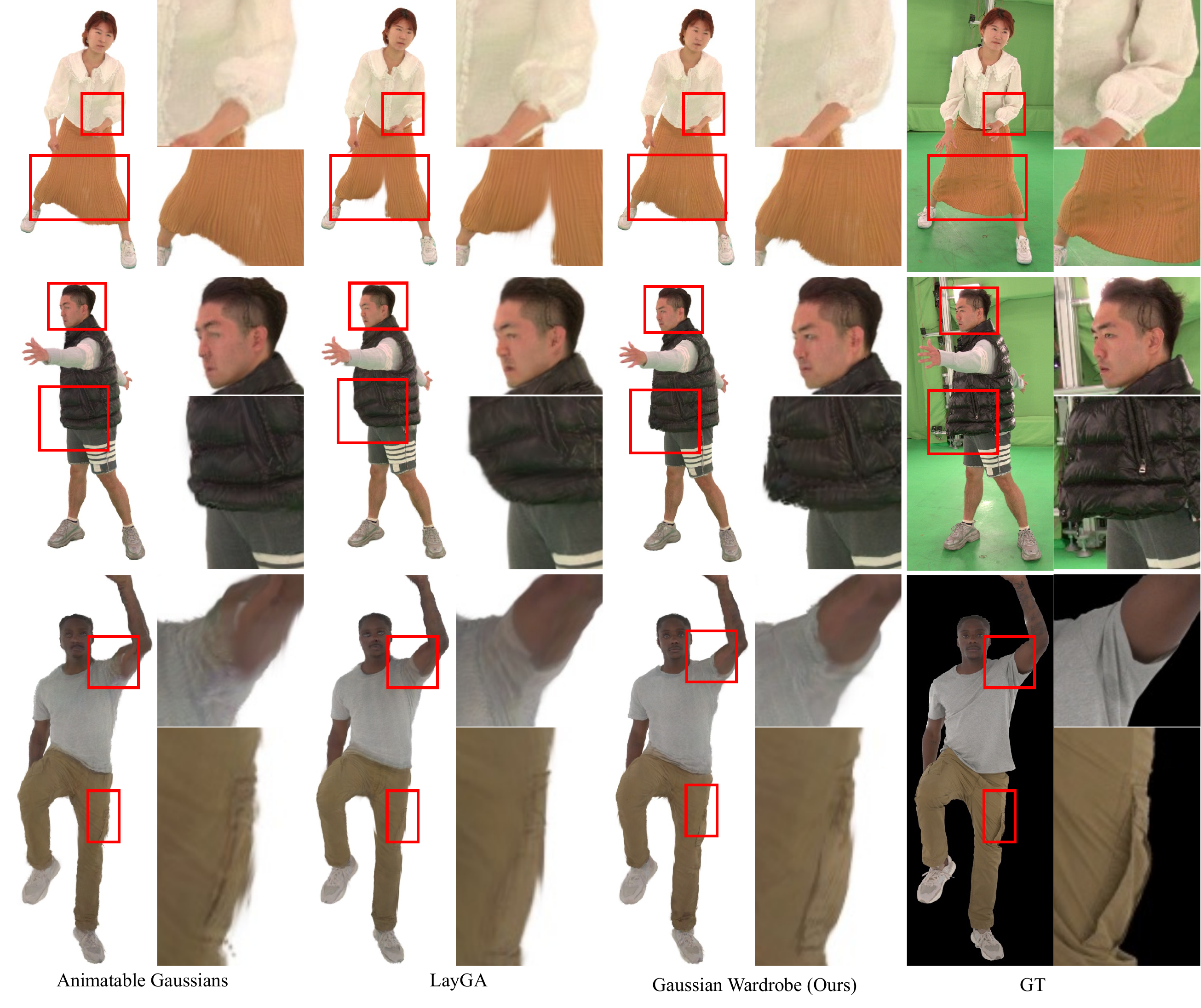}
    \caption{\textbf{Qualitative comparisons of novel pose synthesis on 4D-Dress and ActorsHQ datasets.} Our method can better model the dynamics of free-form garments, such as skirts (top) and vests (middle), and generate realistic renderings with sharper facial and garment details. In contrast, the baseline methods suffer from artifacts, such as blurry faces and semi-transparent clothing, and fail to reproduce fine details like wrinkles or pockets.}
    \label{fig:deform_compare}
    \vspace{1em}

\end{figure*}

%% file: tab/ablation.tex
\begin{table}[t]
    \centering
\begin{tabular}{l|ccc}
\toprule
Method & mIoU $\uparrow$ & Recall $\uparrow$ & F1-Score $\uparrow$  \\
\midrule
w/o $\mathcal{L}_{sg}$ & 0.848 & 0.898 &  0.924 \\
w/o $\mathcal{L}_{reg}$ & 0.879 & 0.922 & 0.940  \\
w/o $\mathcal{L}_{pe}$ & 0.883  & 0.927 & 0.942  \\
Ours Full & \textbf{0.893} & \textbf{0.936} & \textbf{0.947}  \\
\bottomrule
\end{tabular}
\caption{\textbf{Ablation study of the loss terms.} We assess the effectiveness of each loss term from the full objective. We rendered segmentation masks of garment layers and compared them against the ground-truth masks (\cf \cref{fig:ablation-penetration}). The results show that the full model achieves the best performance.}
\label{tab:ablation-regularizers-S}
\end{table}

%% file: sec/FIG_ablation.tex
\begin{figure}[t]
    \centering
    \includegraphics[width=\linewidth]{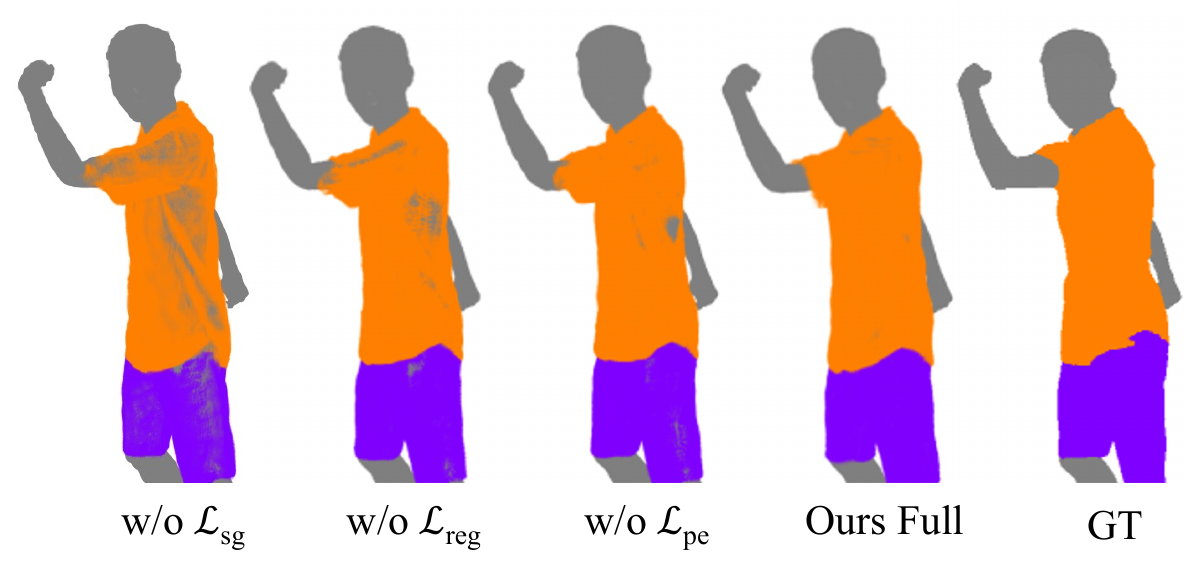}
    \caption{ \textbf{Visualization on the effectiveness of loss terms.}
    We rendered segmentation masks of garment layers to visualize the impact of each regularization term. The evaluation shows that our full model produces cleanest results, while the absence of $\mathcal{L}_{\text{pe}}$, $\mathcal{L}_{\text{reg}}$ and $\mathcal{L}_{sg}$ leads to self-penetration or irregular segmentation.
    }
    \label{fig:ablation-penetration}
\end{figure}

%% file: sec/FIG_segmentation.tex
\begin{figure}[t]
    \centering
    \includegraphics[width=\linewidth]{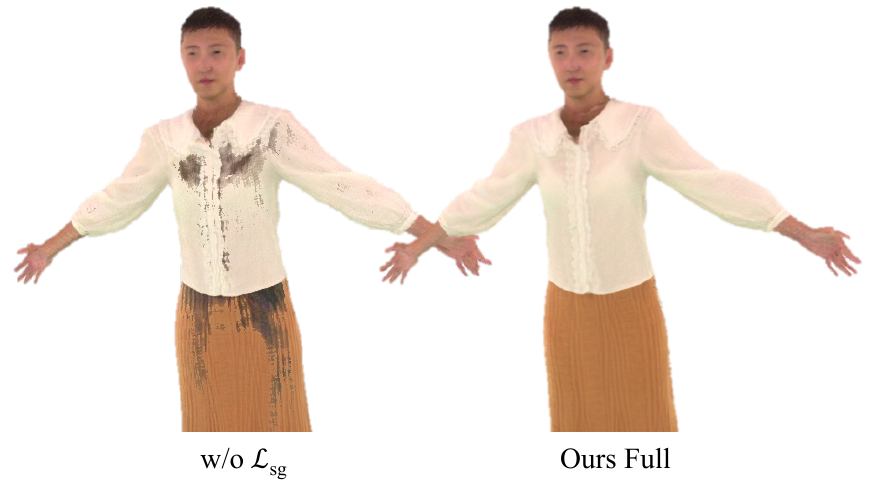}
    \caption{ \textbf{Importance of segmentation loss in virtual try-on.} We visualize the results of virtual try-on using the variants our method trained with and without $\mathcal{L}_{sg}$. The garment’s appearance becomes corrupted due to the entanglement of body and clothing. 
    }
    \label{fig:seg}
\end{figure}

%% file: sec/FIG_virtual_tryon.tex
\begin{figure}[t]
    \centering
    \includegraphics[width=\linewidth]{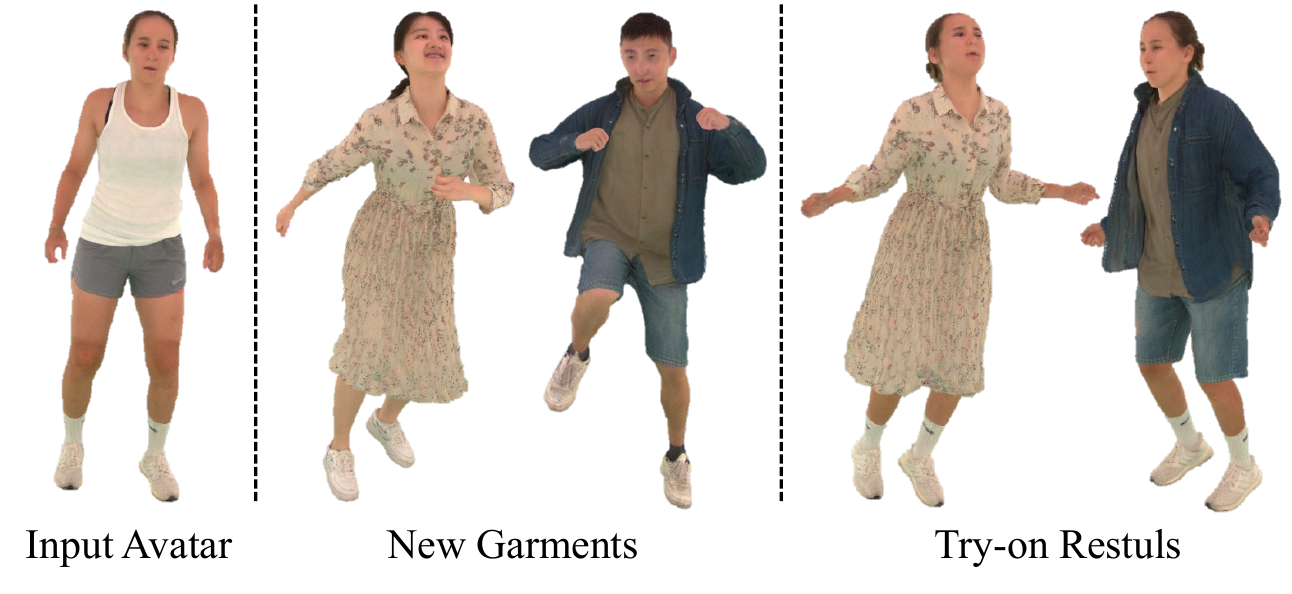}
    \caption{ \textbf{Virtual try-on application.} Our method enables flexible 3D virtual try-on across different subjects. The resulting avatars can be animated to a novel pose.
    }
    \label{fig:clothing_transfer}
\end{figure}

%% file: sec/5_conclusion.tex
\section{Conclusion}

\paragraph{Conclusion.}
We introduced Gaussian Wardrobe, a novel framework for creating compositional 3D neural avatars.
By representing garments as distinct, shape-agnostic 3D Gaussian models, our method enables flexible transfer of clothing across different subjects.
Through extensive quantitative and qualitative experiments, we demonstrated state-of-the-art performance in both reconstruction and animation. We then showcased a virtual try-on application to validate practical relevance. 
In summary, Gaussian Wardrobe offers a promising paradigm for scalable, personalized, and dynamic digital wardrobes. We believe this work paves the way for the future of XR and virtual interaction.

\paragraph{Acknowledgements.} This work was partially supported by the Swiss SERI Consolidation Grant "AI-PERCEIVE".

%% file: supp/6_implement.tex
\clearpage
\setcounter{page}{1}
\maketitlesupplementary
\input{supp/FIG_res_penetration}

\section{Implementation Details}

\paragraph{Normal Estimation for Penetration losses.}
Unlike meshes, Gaussians do not have predefined normals, which makes computing the penetration loss non-trivial.
To estimate normals (i.e., $\mathbf{n}_b$ in \cref{eq:penetration}), we first reconstruct local faces around each Gaussian.
For a Gaussian $b$ at position $\mathbf{p}_b$, we form triangular faces with its counter-clockwise ordered neighbors $\mathcal{N}(b) = \{n_1, ...,n_k\}$. 
We compute the corresponding face normals, sum them, and normalize the result to obtain the final Gaussian normal $\mathbf{n}_b$:

\begin{align}
    \mathbf{N}_b = \sum_{c=1}^{k} (\mathbf{p}_{n_c} - \mathbf{p}_b) &\times (\mathbf{p}_{n_{c+1}} - \mathbf{p}_b) \quad \text{where } n_{k+1} = n_1 \nonumber \\
    \mathbf{n}_b &= \frac{\mathbf{N}_b}{\|\mathbf{N}_b\|}.
\label{eq:normal_estimation}
\end{align}
Specifically, we use $k=4$ neighboring points for the normal calculation. 

\input{supp/TAB_novel_view}
\input{supp/TAB_full_res}

\paragraph{Implementation details of penetration-aware rendering.}

As described in \cref{sec:penetration-aware-rendering}, we use a contour-finding algorithm~\cite{SUZUKI198532} on the segmentation mask $\hat{S}$ of each layer to locate potential penetrations.
This algorithm identifies connected components where pixels classified as an inner layer (e.g., body) are fully enclosed by pixels classified as an outer garment layer (e.g., T-shirt).
To verify whether such regions correspond to true penetrations, we leverage the rendered depth maps of the inner layer ($D_{in}$) and outer layer ($D_{out}$).
For each potential pixel $i$, if $D_{out}[i] - \epsilon < D_{in}[i]$, where $\epsilon$ is a small distance, we classify pixel $i$ as a penetration.
As shown in \cref{fig:penetration_supp}, this method correctly identifies penetrations between the inner T-shirt and the outer jacket.  

\paragraph{Body Gaussians Swapping.}
During virtual try-on, we also swap the offset and rotation parameters of the 3D Gaussians in the body layer.
Taking \cref{fig:penetration} as an example, when replacing the avatar's shorts  $(\mathcal{M}^*_\ell,\mathcal{F}^*_\ell)$ with a new skirt $(\mathcal{M}'_\ell,\mathcal{F}'_\ell)$,  we also update the associated body Gaussians by substituting the parameters $\Delta \mathbf{q}^{c*}$ and $\mathbf{q}^*$ from the original body model $(\mathcal{G}^*_b)$ with the new parameters $\Delta \mathbf{p}^{c'} $ and $\mathbf{q}'$ from the new body model  $(\mathcal{G}'_{b})$.
This swapping process is performed only for the 3D Gaussians located inside the garments, and we found that this strategy mitigates potential penetrations. 
We note that the colors $\mathbf{c}^*$, opacity $\alpha^*$, and scales $\mathbf{s}^*$ remain unchanged to ensure a consistent skin tone during a try-on session.

\paragraph{Network Architecture.}
Inspired by Animatable Gaussians \cite{li2024animatable}, our avatar representation employs a StyleUNet \cite{wang2023styleavatar} variant with two decoders to generate Gaussian maps $\mathbf{M}_L$for both front and back views.
For each body or garment layer, we utilize three separate StyleUNets: one to predict color, one for offsets, and one for the remaining Gaussian attributes.
The input position map $\mathbf{P}_L$ has a resolution of $512 \times 512$, while all output Gaussian maps are produced at $1024 \times 1024 \times 14$. The StyleUNets for color, offsets, and other Gaussian attributes contribute 3, 3, and 8 channels, respectively.
We also adopt the view-dependent color adjustment from Animatable Gaussians \cite{li2024animatable} to model view-dependent effects.

\paragraph{Hyperparameter.}
For photometric loss terms, we use $\lambda_{1} = 0.05$, $\lambda_{2} = 0.1$, and $\lambda_{f} = 0.1$. The segmentation losses are weighted by $\lambda_{sg} = 0.5$ and $\lambda_{bs} = 0.05$. For regularization, we employ $\lambda_{pe} = 0.5$, $\lambda_{o} = 0.005$, $\lambda_{sm} = 0.005$, and $\lambda_{bo} = 0.01$. Additionally, for subjects that include an outer garment (e.g. jackets), we add an extra penetration loss term $\mathcal{L}_\text{pe}$ between $\mathcal{G}_\text{o}$ and $\mathcal{G}_\text{u}$ with the same coefficient $\lambda_\text{pe}$.

\paragraph{Training.}
The StyleUNet models are trained with the Adam optimizer~\cite{adam}, using an initial learning rate of $1.5 \times 10^{-4}$, batch size 1, and 300k iterations. We set $\beta_1 = 0.9$ and $\beta_2 = 0.999$. The learning rate is scheduled with cosine annealing~\cite{SGDR}, decaying gradually to a minimum of $7.5 \times 10^{-6}$.
In addition, to ensure training stability, we apply gradient clipping with a threshold of $5 \times 10^{-4}$. 

Our training process comprises two sequential stages in a coarse-to-fine manner: The first stage is dedicated to reconstructing a coarse shape of the avatar, where the loss weights $\lambda_\text{f}$, $\lambda_\text{sm}$, and $\lambda_\text{pe}$ are set to zero. 
In the second stage, we focus on capturing finer details and animation fidelity with the full training losses. 

The entire training process takes approximately 2.5 days on an NVIDIA RTX 6000 with 24GB VRAM. For the 4D-DRESS subject~\cite{wang20244ddress} with an additional outer clothing, we freeze the learned body network $\mathcal{F}_b$ and jointly finetune $\mathcal{F}_u, \mathcal{F}_\ell, \mathcal{F}_o$ with additional 1.5 days.  

\paragraph{Inference Speed.}
Our method renders avatars with two layers of clothing at 1.08 FPS and three layers at 0.8 FPS on our test hardware. For reference, the baseline Animatable Gaussians~\cite{li2024animatable} renders at 1.5 FPS with the same hardware configurations.

%% file: supp/FIG_res_penetration.tex
\begin{figure}[t]
    \centering
    \includegraphics[width=\linewidth]{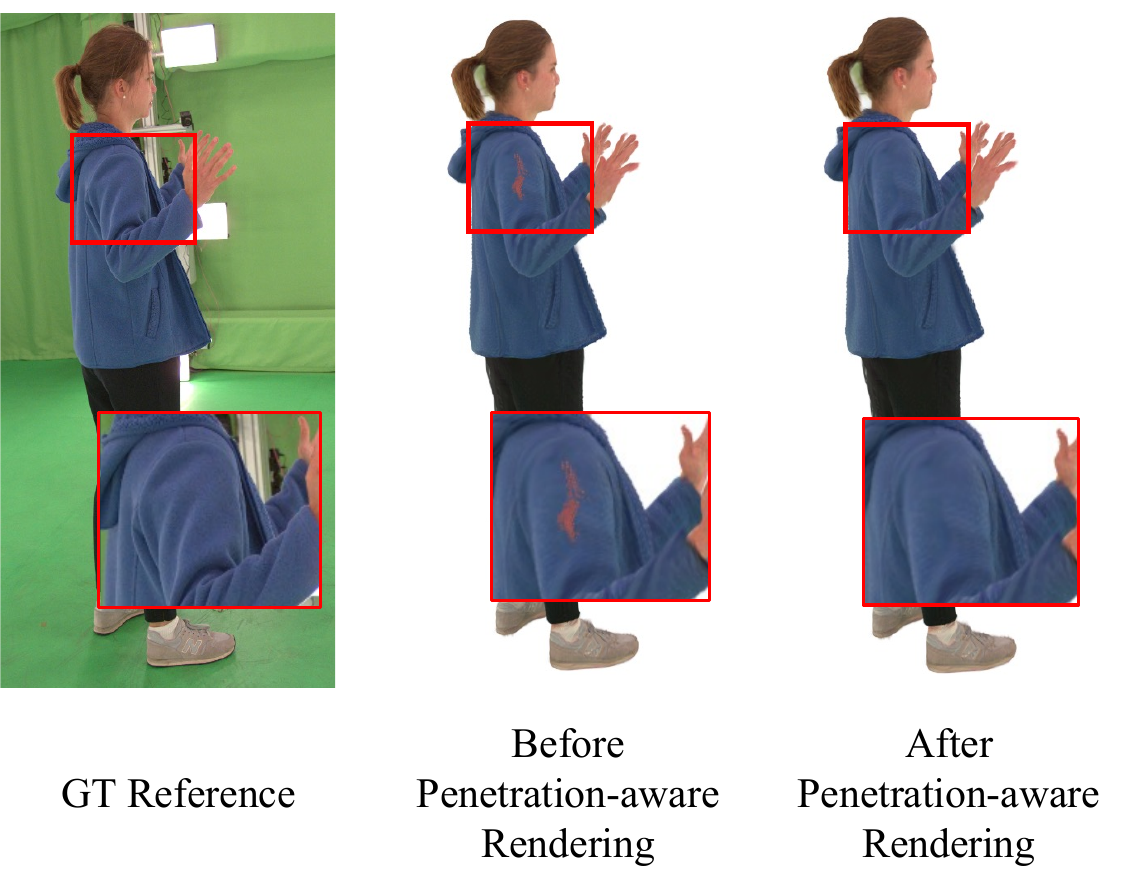}
    \caption{ \textbf{Visualization of penetration-aware rendering.} Penetration-aware rendering removes small artifacts between garment layers.
    }
    \label{fig:penetration_supp}
\end{figure}

%% file: supp/TAB_novel_view.tex
\begin{table}[t]
    \centering
    \small
\begin{tabular}{l|ccc}
\toprule
Method               & PSNR $\uparrow$ & SSIM $\uparrow$ & LPIPS $\downarrow$  \\
\midrule
Animatable Gaussians~\cite{li2024animatable} & 30.51 & \textbf{0.9491} & 0.0332  \\
LayGA~\cite{layga} & 30.36 & 0.9482 & 0.0337  \\
Gaussian Wardrobe (Ours) & \textbf{30.86} & 0.9489 & \textbf{0.0331} \\
 
\bottomrule
\end{tabular}

\caption{\textbf{Quantitative evaluation of novel view synthesis on ActorsHQ dataset.} We use a held-out camera view to evaluate the performance of novel view synthesis.}

\label{tab:novel-view}
\end{table}

%% file: supp/TAB_full_res.tex
\begin{table*}
\centering
\begin{tabular}{l|ccc|ccc|ccc}
\toprule \multirow{2}{*}{Method} & \multicolumn{3}{|c|}{4D-DRESS-00127-Inner} & \multicolumn{3}{|c|}{4D-DRESS-00185-Inner} & \multicolumn{3}{|c}{4D-DRESS-00127-Outer} \\
\cmidrule(r){2-10}& SSIM $\uparrow$ & PSNR $\uparrow$ & LPIPS $\downarrow$ & SSIM $\uparrow$ & PSNR $\uparrow$ &  LPIPS $\downarrow$ & SSIM $\uparrow$ & PSNR $\uparrow$ & LPIPS $\downarrow$ \\
\midrule 
AG~\cite{li2024animatable} &  0.9579 & 28.73 & 0.0503 & 0.9612  & 28.78 & 0.0485 & \textbf{0.9539} &25.62 & 0.0609 \\
LayGA~\cite{layga} & 0.9579 & 28.63 & 0.0506 & 0.9607 & 28.53 & 0.0502 & 0.9533 & 25.49  & 0.0626 \\
Ours & \textbf{0.9585} & \textbf{28.98} & \textbf{0.0502} & \textbf{0.9615} & \textbf{29.15} & \textbf{0.0478} & 0.9535 & \textbf{25.95} & \textbf{0.0604} \\
\toprule
\toprule

\multirow{2}{*}{Method} & \multicolumn{3}{|c|}{ActorsHQ-Actor05} & \multicolumn{3}{|c|}{ActorsHQ-Actor08} & \multicolumn{3}{|c}{ActorsHQ-Actor01} \\
\cmidrule(r){2-10}& SSIM $\uparrow$ & PSNR $\uparrow$ & LPIPS $\downarrow$ & SSIM $\uparrow$ & PSNR $\uparrow$ &  LPIPS  $\downarrow$ & SSIM $\uparrow$ & PSNR $\uparrow$ & LPIPS $\downarrow$ \\
\midrule 
AG~\cite{li2024animatable} &  0.9480 & 28.98 & 0.0387 & 0.9321 & 27.62 & 0.0474 &0.9454  & 27.22& 0.0380\\
LayGA~\cite{layga} & 0.9459 & 28.44 & 0.0402 & 0.9342 & 27.48  & 0.0465 & 0.9456 & 27.55& 0.0384 \\
Ours & \textbf{0.9484} & \textbf{29.29} & \textbf{0.0381} & \textbf{0.9361}& \textbf{27.99} & \textbf{0.0455}  & \textbf{0.9482} & \textbf{27.97} & \textbf{0.0341}  \\

\bottomrule
\end{tabular}
\caption{\textbf{Full quantitative evaluation of novel pose synthesis on the 4D-DRESS and ActorsHQ dataset.}}
\label{tab:full_novelpose}
\end{table*}

%% file: supp/7_more_results.tex
\input{supp/FIG_canon}

\section{More Experimental Results}

\subsection{Additional Novel View Synthesis Results}
Although not the primary focus of our work, we also evaluated novel view synthesis using the ActorsHQ dataset \cite{isik2023humanrf}.
We used a held-out camera view from the training videos and computed the average image metrics across the three subjects.
The qualitative and quantitative results, presented in~\cref{fig:novelview2},~\cref{fig:novelview1}, and~\cref{tab:novel-view}, demonstrate that our method achieves photorealistic rendering under novel views.

\subsection{Full Results on Novel Pose Synthesis}
In our main paper, we reported the average metrics per dataset. For completeness, \cref{tab:full_novelpose} presents the full quantitative results for each subject on the novel-pose synthesis task.

\subsection{Diffused Skinning Fields}
As described in \cref{sec:rep}, our method first deforms the template into the canonical space using inverse LBS. For this step, we adopted a diffused skinning field strategy \cite{Chen2023fastsnarf}. In \cref{fig:canon}, we compare the quality of different techniques for querying skinning weights in inverse LBS.

\subsection{More Virtual Try-On}
We show more virtual try-on results in \cref{fig:supp-more_tryon}. Please also refer to our video for more visual results.

\input{supp/FIG_try-on}
\input{supp/FIG_novelview}

\section{Discussion and Future Work}
\paragraph{Scaling up to monocular videos.} Gaussian Wardrobe is currently implemented and trained with multi-view videos captured in the lab environments, including accurate 3D poses and segmentation masks. However, such a data requirement is a bottleneck to scaling up the method for broader use base. Our ultimate goal is to adapt our method to monocular videos captured by modern smartphones in daily life. A promising solution is to integrate personalized pose tracking~\cite{ho2025phd} with video-based reconstruction methods~\cite{paudel2024ihuman,reloo} when ground-truth poses are not available. 

\paragraph{Virtual bone-based deformations.} While our method can handle loose garment deformations within the training pose distribution, modeling deformations for out-of-distribution poses is still challenging. Thus, an exciting direction for future work is to extend our layered representation with virtual bones~\cite{reloo,tan2025dressrecon} to achieve higher-fidelity garment modeling.

%% file: supp/FIG_canon.tex
\begin{figure*}[t]
    \centering
    \includegraphics[width=\linewidth]{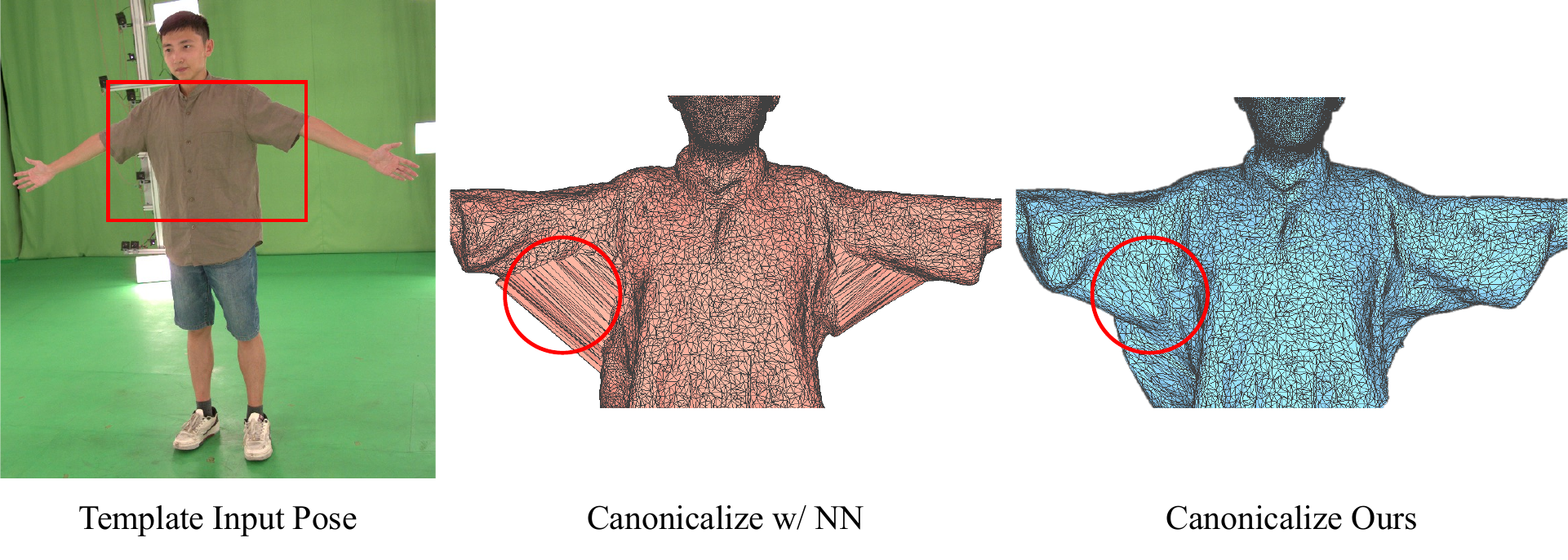}
    \caption{
    \textbf{Importance of diffused skinning fields.} 
    To deform the template mesh into the canonical pose, we evaluate different strategies for querying skinning weights in inverse LBS. Nearest-neighbor querying fails to preserve topology near the armpits, whereas our diffused skinning fields produce smoother and more accurate garment meshes.
    }
    \label{fig:canon}
\end{figure*}

%% file: supp/FIG_try-on.tex
\begin{figure*}[t]
    \centering
    \includegraphics[width=0.90\linewidth]{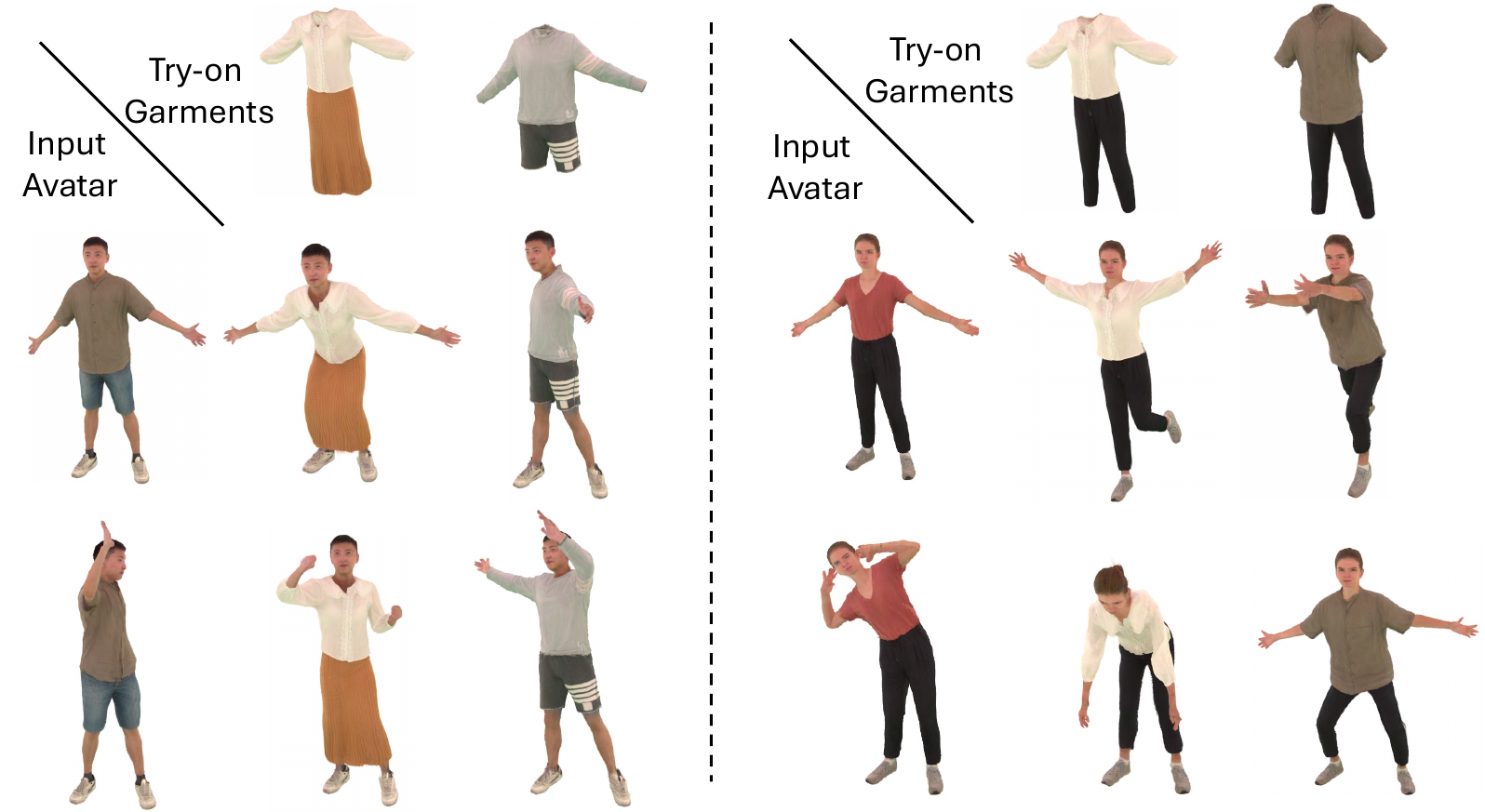}
    \caption{\textbf{More results on virtual try-on.} Our method supports flexible 3D try-on across different subjects, and the generated avatars can be animated into novel poses.
    }
    \label{fig:supp-more_tryon}
\end{figure*}

%% file: supp/FIG_novelview.tex
\begin{figure*}[t]
    \centering
    \includegraphics[width=\linewidth]{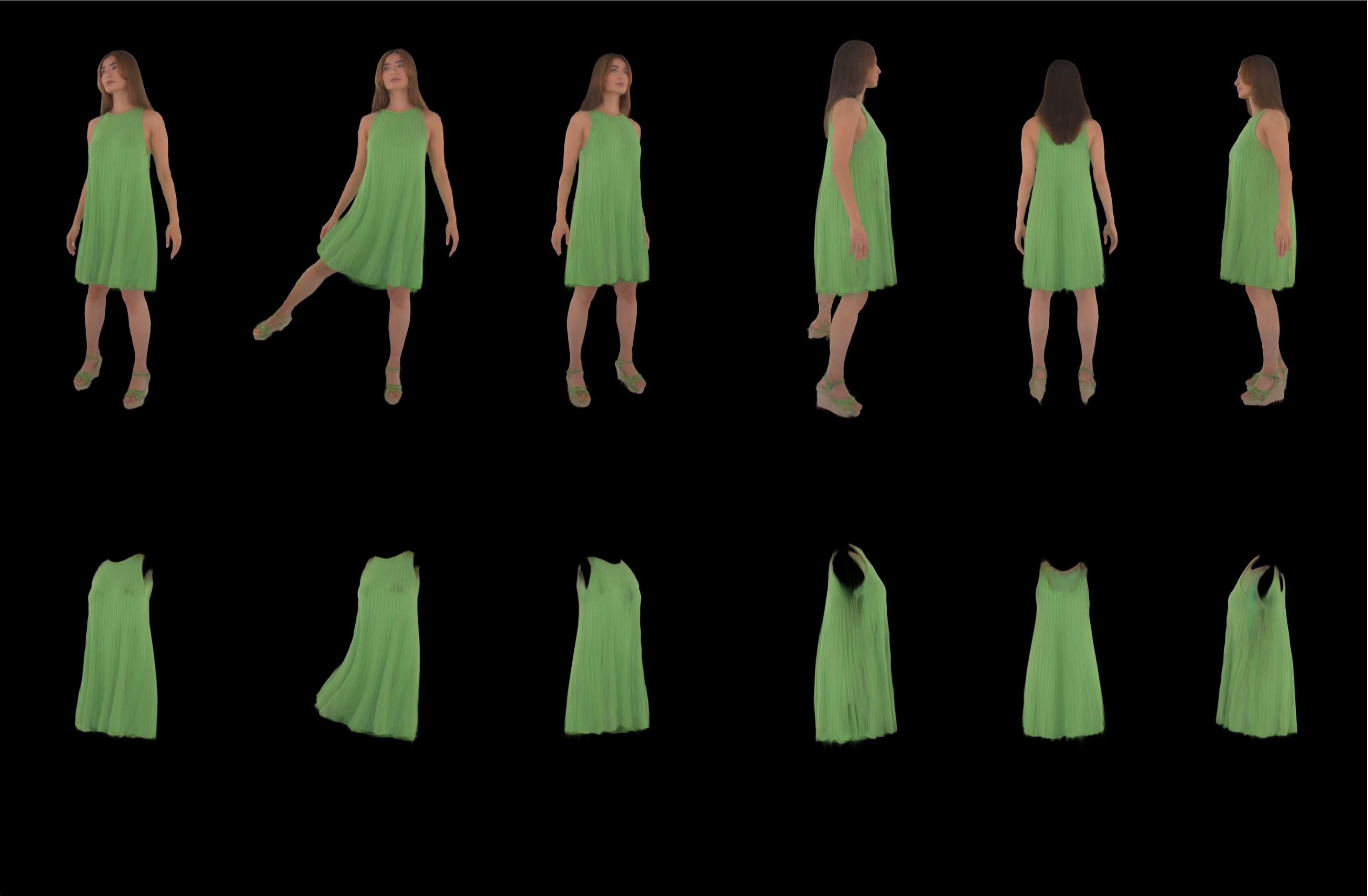}
    \caption{\textbf{Results on ActorsHQ.} Our method produces photorealistic 360{\textdegree} rendering under novel views and poses.
    }
    \label{fig:novelview2}
\end{figure*}

\begin{figure*}[t]
    \centering
    \includegraphics[width=0.90\linewidth]{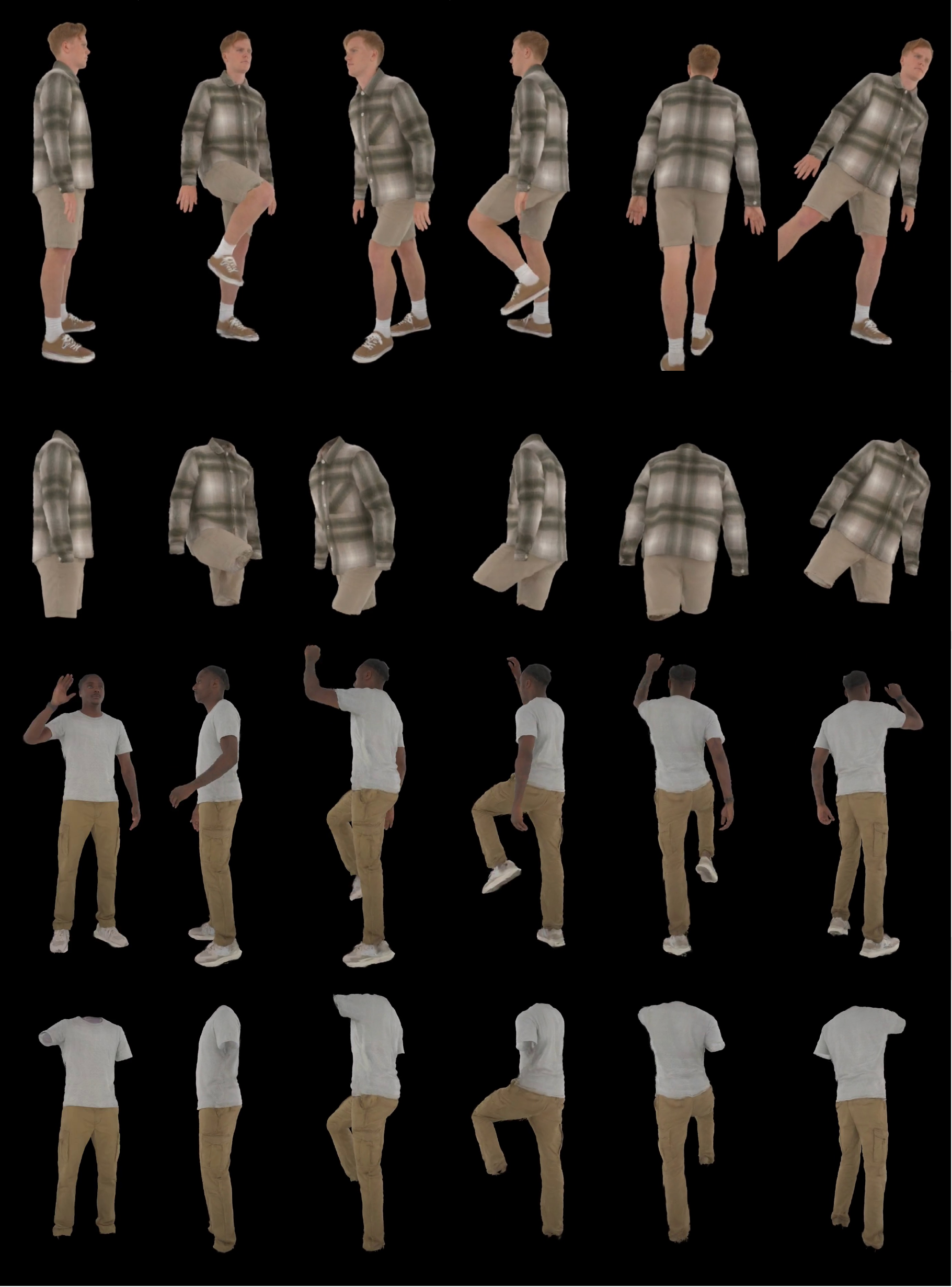}
    \caption{\textbf{Results on ActorsHQ.} Our method produces photorealistic 360{\textdegree} rendering under novel views and poses.
    }
    \label{fig:novelview1}
\end{figure*}